# Robustness of different loss functions and their impact on network's learning capability

Vishal Rajput[1,*]

[1]Computer Science Department, KU Leuven, Belgium

[*]Corresponding author. Email: vishal.rajput@student.kuleuven.be

**Abstract** Recent developments in AI have made it ubiquitous, every industry is trying to adopt some form of intelligent processing of their data. Despite so many advances in the field, AI's full capability is yet to be exploited by the industry. Industries that involve some risk factors still remain cautious about the usage of AI due to the lack of trust in such autonomous systems. Present-day AI might be very good in a lot of things but it is very bad in reasoning and this behavior of AI can lead to catastrophic results. Autonomous cars crashing into a person or a drone getting stuck in a tree are a few examples where AI decisions lead to catastrophic results. To develop insight and generate an explanation about the learning capability of AI, we will try to analyze the working of loss functions. For our case, we will use two sets of loss functions, generalized loss functions like Binary cross-entropy or BCE and specialized loss functions like Dice loss or focal loss. Through a series of experiments, we will establish whether combining different loss functions is better than using a single loss function and if yes, then what is the reason behind it. In order to establish the difference between generalized loss and specialized losses, we will train several models using the above-mentioned losses and then compare their robustness on adversarial examples. In particular, we will look at how fast the accuracy of different models decreases when we change the pixels corresponding to the most salient gradients.

## 1 Introduction

For a decade or so people are using deep learning and Artificial intelligence tools in a variety of places. Despite being such a powerful tool, it is still missing from a lot of industries. In most of the places where there is some risk involved, usage of AI remains less because of the trust issues humans have with past autonomous systems. Present-day AI might be very good in a lot of things but it's very bad in reasoning and this behavior of AI can lead to catastrophic results. Over the past few years, we have seen some incredible feats achieved by AI, but at the same time, it has failed badly in areas like autonomous drive. Current AI systems are very good at capturing patterns hidden in a myriad of data streams but sometimes it fails even for the simplest of the test case. If a bank rejects and accepts loan requests based on the decision of their AI model which they don't understand fully, it can lead to huge monetary losses. Banks need to give a valid and acceptable reason to their customers as to why their loan application was rejected. For medical purposes, it is straightforward a matter of life and death. The above mentioned are few areas where explainability for AI's decision is a must.

Although rigorous research has been going on in the mathematical explanation of these models, the question still remains the same how to make general people have trust in AI. Explaining a model's working and the decision to an expert is not going to make it acceptable to the laymen. In 2015, Nguyen [1] showed how easy it is to fool a convolutional neural network into making an erroneous classification with very high confidence. The predicted objects were not even remotely close to their real counterparts. In [2] researcher talked about trust and transparency in AI systems. In many cases, it has been shown that the network looks at completely different objects while classifying different objects. An autonomous vehicle hitting people, random noise being predicted as different objects with high accuracy are few examples as to why we can't trust AI in critical situations. In the case of classical machine learning, decision trees are considered to be the best method for generating explanations. Decision trees result can be easily transformed 1 into human interpretable if and else statements. Contrary to Machine Learning algorithms, Deep learning has several million parameters, so understanding the decision of Deep learning is a bit tricky. There are two major ways in which DL models can be interpreted, first one is to generate some sentences with each prediction in human-readable language or produce some heatmaps as to where the model is looking at while making its decision. Explainability in AI leads to Responsible Artificial Intelligence, namely, a methodology for the large-scale implementation of AI solutions in real-life scenarios with fairness, explainability and accountability at its core.

## 2 Related Works

In 2016, researchers provided a brief discussion about what can be visualized over the whole training process. They presented a premise of visualizing the data before implementing any particular model because even with a slight change in the dataset, a huge change might occur in the learned features. During training, plotting its criterion, the learning rate, test accuracy on the validation set tells us a lot about whether the network is being trained in the right direction or not. This is a good start but nowhere close to explain the AI's decision. In the work presented by Ribiero et al. [3], it was shown that while classifying between the wolves and huskies, the network was giving more than 90 percent accuracy but it was not even looking at the subject matter to make its decision. To prevent such catastrophe's researchers came up with the idea of visualizing the gradients in order to assess the importance of different regions of the image. Over the years, several methodologies were developed like DeepLift [4], it compares the activation of each neuron with a corresponding reference activation, and according to the difference, it evaluates the contribution of the neuron to the network's decision. Their idea was based on the fact that a low activation value doesn't mean that a neuron is less significant. However, comparing both activations from a reference input and the given input can reveal the importance of a feature. Few other common methods for gradient visualization include Gradient Input callback [5] SmoothGrad [6], Integrated gradients [7]. Another very interesting idea about visualization comes from Zeiler et al. [8]. It basically tries to visualize how different part of the image affects the results of a neural network, we hide or occlude certain parts of the image and then visualize the results of the network. The areas or sections of the image that are more important will lead to a larger drop in the accuracy, and that's how we form our saliency maps [9].

In 2017, Selvaraju et al. [10] came up with the Grad-CAM, it calculates the gradient of a particular convolutional layer with respect to the input image. Grad-CAM uses the information flowing in the form of gradients to the last convolutional layer to understand each neuron's decision. In this method, it calculates the gradient of the score of the class C with respect to the feature maps to obtain the class discriminative localization map. The flowing gradients are passed through global average pooling to obtain the importance of the neuron for the class C. After this step, a weighted combination of activation maps is calculated followed by a ReLu activation function. The result obtained is a coarse heatmap of the same size as of that particular convolution layer. Grad-CAM visualizations are very good, they can localize accurately and also class discriminative but they lack the ability to show fine-grained results like the ones given by the Guided Backpropagation [11]. In order to solve this issue researchers combined Grad-CAM and Guided Backpropagation to generate pixel level, class discriminative heatmaps.

In past, Szegedy et al. [12] have shown that for every deep learning model we can generate adversarial examples that can fool any network. It takes much more effort to fool a well-trained and generalized network compared to a badly trained network. In 2014, Goodfellow et al. [13] reasoned why adversarial examples exist. Goodfellow et al. believed that adversarial examples exists because models are too linear. Adversarial perturbations are dependent on model's weights, which are similar for different models learned to perform the same task. A generalization of adversarial noise was observed across different natural examples, adversarial perturbations are not dependent on a specific point in space but on the direction of the perturbation. They also showed that models which are easy to optimize yield easily to adversarial and fooling examples, thus they have no capacity to resist these perturbations. Keeping the ideas given by Goodfellow et al. in mind we will try to analyze whether a networks generalization capability and robustness against adversarial examples changes with change in the loss function. This work will provide visual insights as to why one loss function performs better than the other in terms of generalizations and robustness. We will try to provide a visual and quantitative explanation as to why combining the different loss functions is a good idea.

## 3 Methodology

For the purpose of this work, we are going to work with medical image segmentation dataset [14] released in 2019 under the medical segmentation decathlon challenge. This dataset consisted of scans of 10 different body parts namely Liver Tumors, Brain Tumors, Hippocampus, Lung Tumor, Prostate, Cardiac, Pancreas Tumor, Colon Cancer, Hepatic Vessels and Spleen. For our purpose tests are performed majorly on Hippocampus data. It had 395 3-D images out of which 263 are for training and 131 for testing. Every image in the dataset had different widths, heights and depths. Training neural networks with different sized input is pretty hard, so we zero-padded the data along width and height and treated each frame along the depth axis as a separate sample. Instead of making 3-D convolutional models, we decided to go with the more traditional 2D conv models. For the task of generating segmentation mask several architectures were tested initially, after comparing the results of all the network we decided to proceed with U-Net architecture. U-Net compresses the incoming inputs and then scales them back to the original size, it is similar to what we have in autoencoders [15], but it also has skip connections from the compression side to the decompression side.

*Entropy Loss:* Entropy is a measure of uncertainity in a random variable X's outcome. The ones we use for the training of networks are logarithmic loss, log loss or logistic loss. Each predicted class probability is comapred to the actual class output, in our case, we have a binary classification thus output is limited to 0 and 1. The penalty is logarithmic in nature and yields value closer to zero when predicted values are close to the real value.

$$Loss_{CE} = \sum_{i=1}^{n} y_i \log(p_i)$$

Here $y_i$ is the real class label and $p_i$ is the predicted value, for our case, we only have two labels 0 and 1, thus cross entropy loss will be called as Binary Cross Entropy or BCE. Another very common loss is called Mean Square Error or MSE, but it has been shown in past by several researchers that BCE is generally used for classification whereas MSE is more suited for regression type of problems.

*Dice Loss:* The Dice score coefficient is a measure of overlap, generally used in image segmentation. The advantage of using dice loss is that it can very well handle the class imbalance in terms of pixel count for foreground and background. Other generalized loss function like MSE or BCE are not great at handling class imbalances, we have weighted cross entropy that somewhat alleviates the class imbalance issue but still it is not as good as Dice loss.

$$Loss_{Dice} = 1 - \frac{2\sum_{i=1}^{n} p_i y_i}{\sum_{i=1}^{n} p_i^2 + \sum_{i=1}^{n} y_i^2}$$

Here $y_i$ is the real pixel value and $p_i$ is the predicted pixel value.

*Focal Loss:* Focal loss is just an extension of the cross-entropy loss function that weighs down the easy examples and give more priority to hard examples during the training of the network. When an example is misclassified and pi is small, the modulating factor is near 1 and the loss remains unaffected, but when $p_i \to 1$, the factor goes to 0, the loss for well-classified examples is down weighed. The focusing parameter $\gamma$ smoothly adjusts the rate at which easy examples are down-weighted.

$$Loss_{Focal} = \alpha_i (1 - p_i)^\gamma \log(p_i)$$

Here $\alpha_i$ is a weighted term whose value is $\alpha$ for positive (foreground) class and $1 - \alpha$ for negative (background) class, $p_i$ is the predicted class probability for a given pixel and $\gamma$ is tunable focusing parameter.

For the purpose of experimentation, we trained four U-net models with different loss functions namely BCE, Dice loss, BCE + Dice loss and BCE + Dice loss + Focal loss and compared their performance both qualitatively and quantitatively. For the qualitative comparison we visualize the gradients and how they are distributed around the segmentation mask. Gradients can be calculated in a lot of ways, i.e., gradient between loss matrix and input or predicted output and input or a particular conv layer and input, for our work we are focusing on the gradient calculated between predicted output and input matrices. Manual inspection of the distribution of gradients will tell us how and why the four models behave differently, it will also give on insight as to which model is more likely to be robust against adversarial examples. Another visual inspection which will be discussed in greater detail in the next section is how does the segmentation mask starts shrinking with adding more perturbations to the input image.

For the quantitative results, we will look at the trend lines of Dice score for all the four models given the different sets of adversarial examples. An important thing to note here is that we have scaled all our images between 0 and 0.2 (for training) because we wanted to see the effect of what happens when the noise level goes beyond the maximum pixel value (maximum pixel value=0.2, noise level=[0, 1]) in an input image and compare those results with perturbations happening at the same scale (maximum pixel value=0.2 and noise level=[0, 0.2]). For testing out the robustness and generalization capabilities of all the four models, first we identify the pixels corresponding to 5-highest gradient and replace those pixels with (i) with maximum pixel value i.e., 0.2 (in our case) (ii) with value of 0, 0.1, 0.2, 0.4, 0.6 and 0.8 and then we see the effect of pixel attacks on N-pixels corresponding to the N-highest gradient value, i.e. we replace N-pixels (i) with 0.2 and (ii) with random values between 0 and 1 (for this case each experiment is run 100 times).

## 3 Results and Discussions

As discusses in the above sections, using gradient information to understand the black-box nature of the DL models is a very good idea and it also helps in understanding the difference in a model's performance with change in the loss function. When we change the loss function, we basically change the optimization problem thus it is necessary to understand those differences and how it relates to the real-world performance. As described in the methodology section, we'll be comparing the gradients for all the four models trained with BCE, Dice loss, BCE + Dice loss and BCE + Dice loss + Focal loss respectively. Comparing the results of all the four model will give us a very good idea about the working of both the specialized loss functions and generalized loss functions. We will also try to understand why we need different sets of loss functions for different problem, because theoretically we can optimize any model using any loss function, but it doesn't work properly in real life scenarios. All the gradient images presented in this section are calculated between the output and the input.

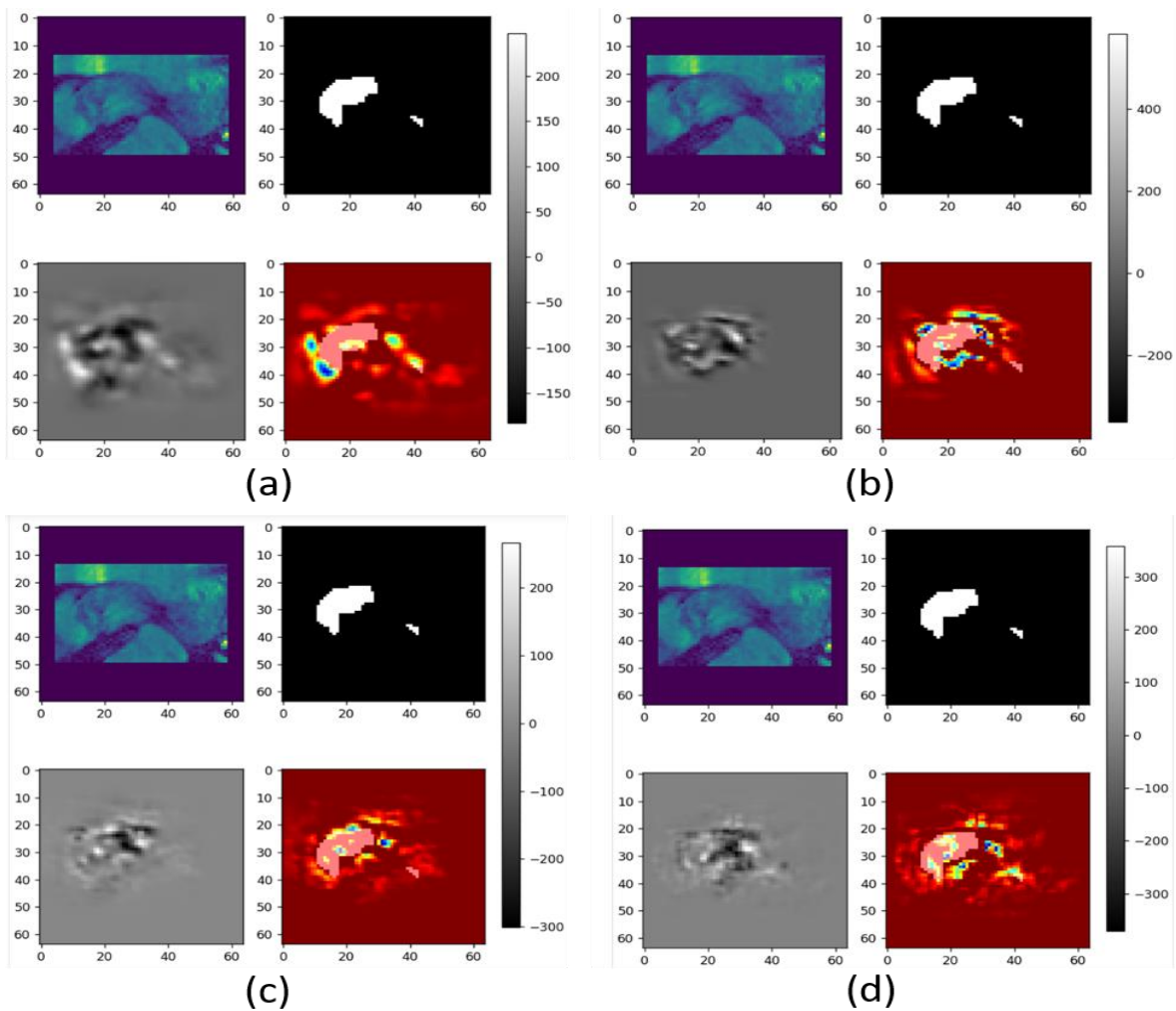

**Figure 1:** Gradient visualization (calculated between input and output image). Original image, original segmentation mask (top row), gradients in grayscale, and gradients overlaid on original segmentation mask (bottom row) for (a) BCE, (b) Dice loss, (c) BCE+ Dice loss, and (d) BCE+ Dice loss + Focal loss

In Figure 1, we see the original image, original segmentation mask, gradients (grayscale, includes both positive and negative gradients) and positive gradients overlaid on top of the original segmentation mask for all the four models. There are two things to be noted from the manual inspection of the gradient image, firstly, the scale of the gradients, secondly, the distribution of the gradient. If we look at the distribution of the gradients for model trained with BCE (a), we see that active gradients have a spread over much larger region compared to the active gradients of model trained with Dice loss (b). Another difference is in the scale of gradients for (a) and (b), (b) has much sharper transition from active to inactive gradients compared to (a). We believe that the robustness of a model against adversarial attacks is highly dependent upon how large is the spread of active gradients. More the spread of active gradients, better it will perform against the adversarial attacks. We also believe that Dice loss focus much harder on few pixels compared to BCE and that's why it gives better performance in terms of Dice Score compared to BCE. We believe that loss function that focuses hardly (i.e., sharp transition in gradients) will generally perform better in terms of accuracy, in our case, Dice score. Now, if we look at (c) and (d), we see that their active gradients are somewhere in between (a) and (b). More focused than (a) and more dispersed than (b). Combining loss functions gives us the combined property of both BCE and Dice, i.e., better accuracy or Dice score of Dice loss and better robustness of BCE.

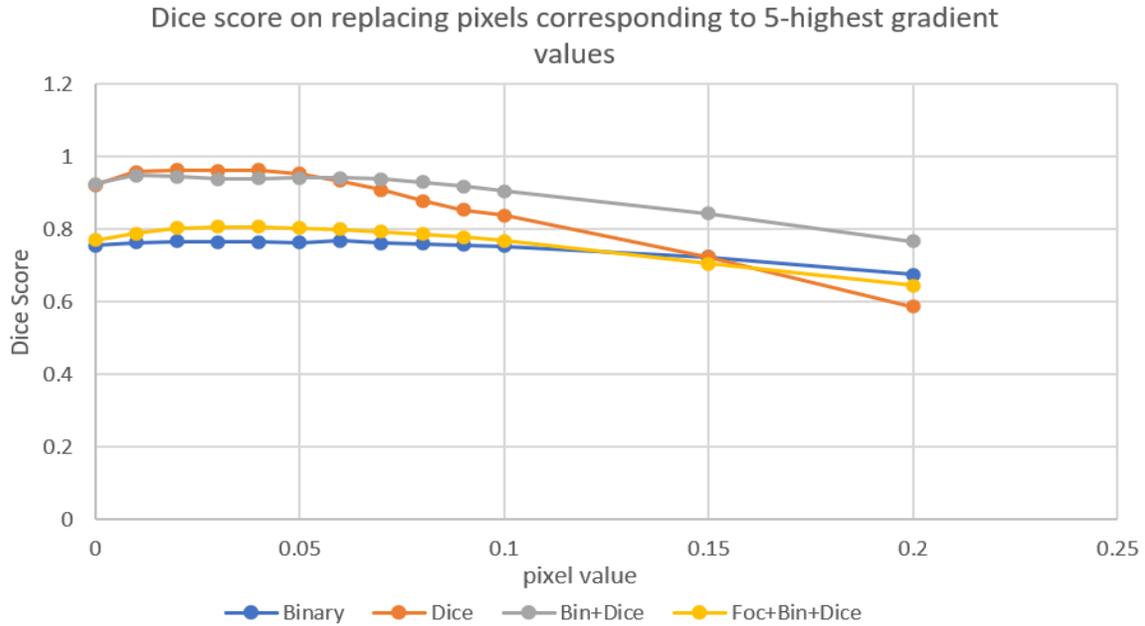

**Figure 2:** Effect of replacing pixels (noise level=[0, 0.2]) corresponding to 5-highest gradient values for model trained with BCE, Dice loss, BCE + Dice loss, and BCE+ Dice loss + Focal loss

To substantiate our claim that spreading active gradients in wider region helps in more robust models we will look at the the results of adversarial attacks. If we look at the graph of Figure 2, we can see that as we replace the pixels corresponding to the 5-highest gradient value, Dice Score starts to reduce. Models trained with Dice loss and BCE+Dice loss starts from the similar Dice score but as the level of perturbation increases model with Dice loss starts decreasing more rapidly compared to the model trained with BCE+Dice loss. We can also see the amount of drop in Dice score is very loss for both BCE and Focal. As mentioned in the above sections, we analyzed the effect of noise on the Dice score when level of noise exceeds the maximum pixel value in an image. Figure 3 shows us how the four models behave for very high level of perturbations. We can see that by the time the perturbation reaches three times the maximum pixel value (0.2 in our case, noise level=0.6) Dice score for the model trained with Dice loss has already dipped below 5% whereas for the model trained with BCE+Dice loss, Dice Score remains around 60%, a huge improvement in terms of robustness. We can also see in Figure 4 how the segmentation mask shrinks with the increased level of noise. For Dice loss (b) by the time noise level reaches to 0.4 segmentation mask becomes pretty useless on the other hand for the model trained with BCE+Dice loss and BCE+Dice loss+Focal loss performs better at noise level of 0.6 compared to the Dice performs at the noise level of 0.4. Another difference that we can see between model trained with BCE and Dice loss is that Dice generates very hard margin whereas BCE generates soft margin

(During the training of the networks, we didn't use softmax at the last layer, to observe the difference in decision boundaries due to the usage of different losses).

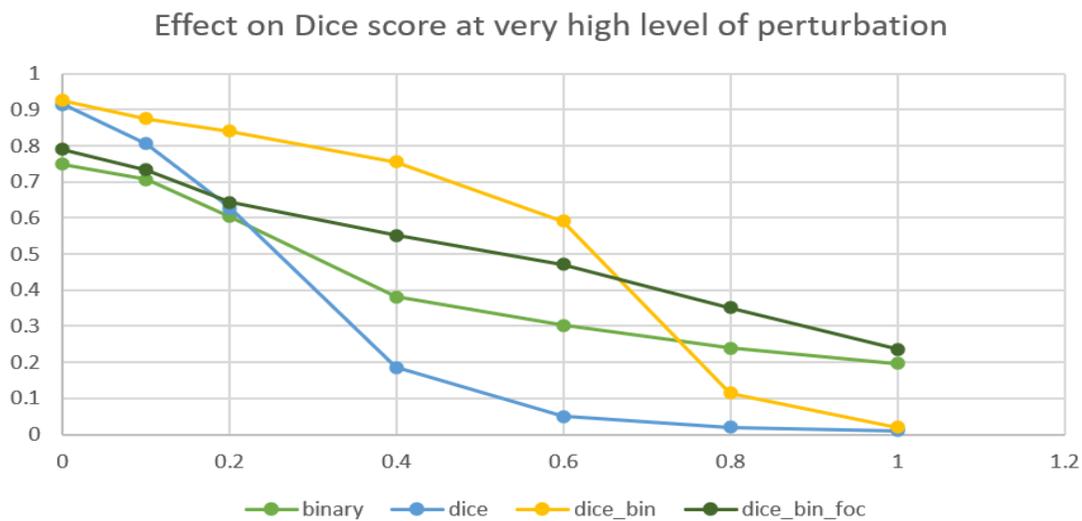

**Figure 3:** Effect of replacing pixels (noise level=[0, 1]) corresponding to 5-highest gradient values for model trained with BCE, Dice loss, BCE + Dice loss, and BCE + Dice loss + Focal loss

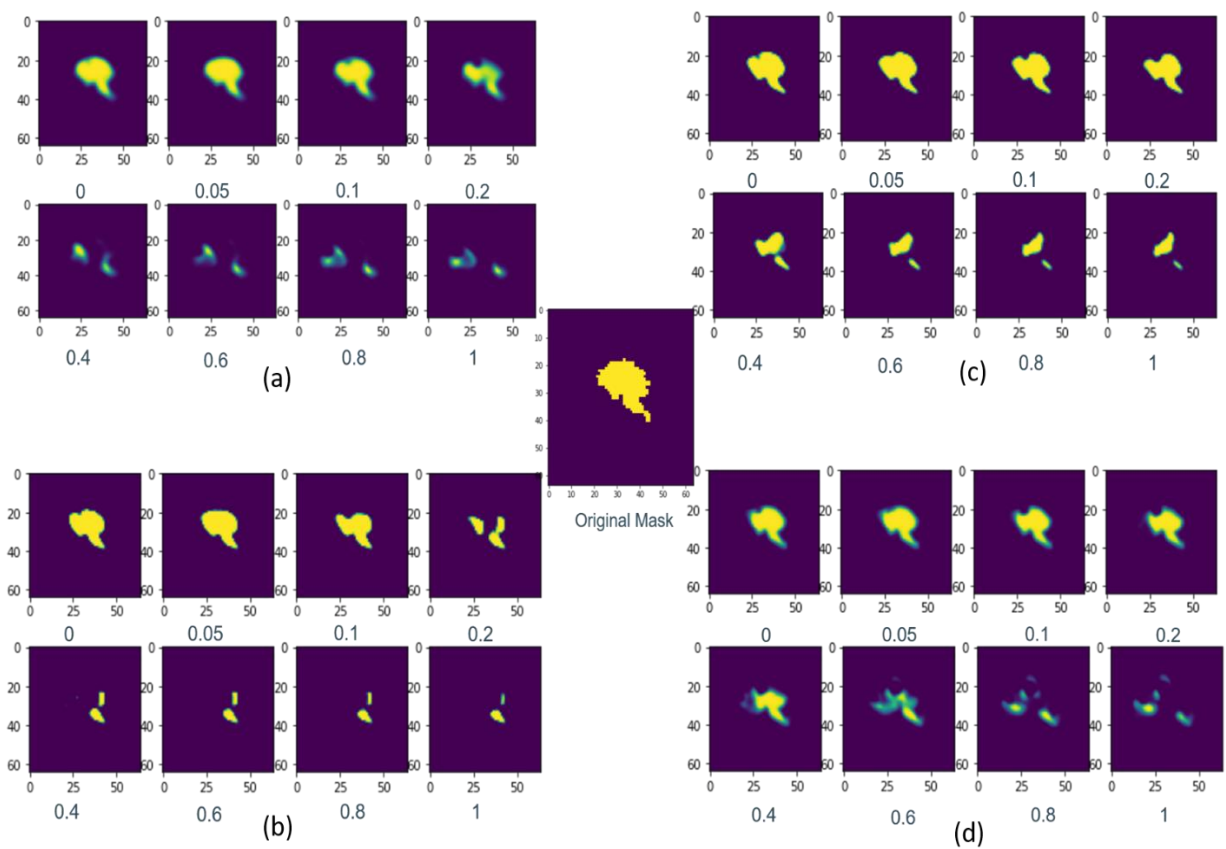

**Figure 4:** Original segmentation mask and shrinking of segmentation mask by replacing pixels (noise level=[0, 1]) corresponding to 5-highest gradient values for model trained with (a) BCE, (b) Dice loss, (c) BCE + Dice loss, and (d) BCE+ Dice loss + Focal loss

We can see from these results that Dice even though gives high accuracy or Dice Score, but it is susceptible to the noise. Let's see the effect on Dice Score for all four models, when we replace pixels corresponding to N-highest gradient instead of replacing only 5 pixels. Figure 5 show us the result of replacing N-pixels with noise level=0.2. Again, we observe the similar trend, by the time we replace 30 pixels by 0.2 Dice score for the model trained with Dice loss has reduced to 30% whereas for the model trained with BCE+Dice loss is still above 70%. Next, we replace N-pixels with noise level=[0, 1], since we are using random value as noise so we need to run this experiment multiple times (we averaged over 100 iterations) in order to properly assess the difference in the Dice score for all the four models. Result of Figure 6 also shows the similar trend i.e., BCE+Dice outperforming Dice in terms of robustness against adversarial attacks. Combining loss functions not only gave us more robust models but we also saw a little increase in the accuracy or Dice Score. We believe that adding combining a generalized loss along with a specialized loss helps in regularization of the network thus outperforming the models trained with individual loss functions both in terms of accuracy and robustness. Figure 7 shows us the result of all the four models without any perturbations, we can see that all the four models generate very similar segmentation masks but their performance against adversarial attacks varies hugely.

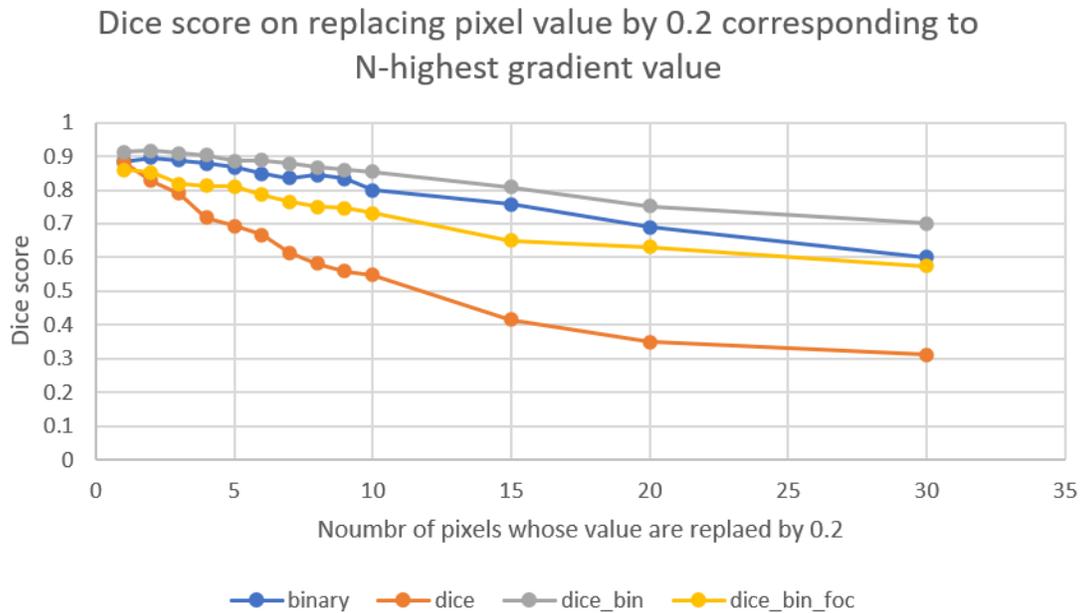

**Figure 5:** Effect of replacing pixels (noise level=0.2) corresponding to N-highest gradient values for model trained with BCE, Dice loss, BCE + Dice loss, and BCE+ Dice loss + Focal loss

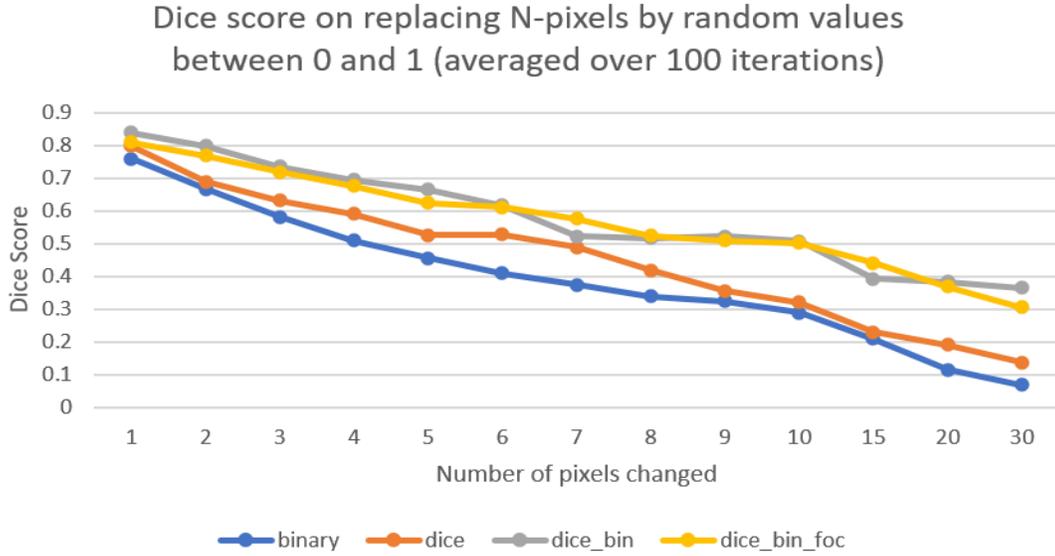

**Figure 6:** Effect of replacing pixels (noise level=[0, 1]) corresponding to N-highest gradient values for model trained with BCE, Dice loss, BCE + Dice loss, and BCE + Dice loss + Focal loss

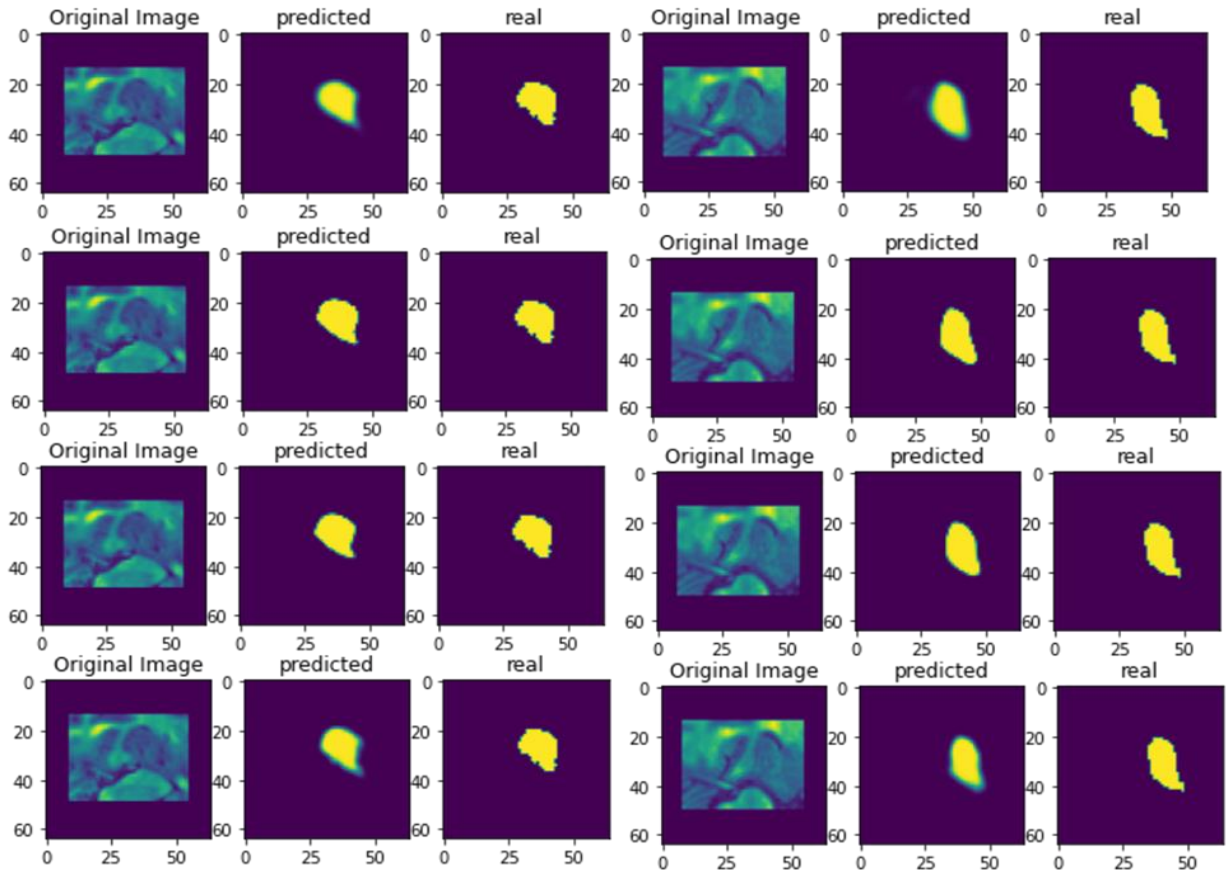

**Figure 7:** Original image, predicted segmentation mask and real segmentation mask for BCE (1st row), Dice loss (2nd row), BCE + Dice (3rd row) and BCE + Dice loss + Focal loss (Fourth row

# 5 Conclusion and Future Works

From the results discussed in the above sections, we can conclude both qualitatively and quantitatively that Dice loss gives us more accurate results for segmentation compared to BCE. We can say with confidence that combining BCE and Dice loss gives us the best result in terms of segmentation, pixel-wise accuracy, generalization and adversarial attacks. We observed that changing a few pixels are enough to render the models useless. It is easier to detect noise in natural images but to detect perturbation in medical images is very hard thus we need to be cautious about the AI's result on medical images. We observe that using specialized loss functions can force the network to focus very hard on very few pixels thus making them susceptible to adversarial attacks. BCE is much more resilient to both the level of noise and the number of pixels changed. Both visual and quantitative results corroborate our initial hypothesis of Dice loss being more accurate and BCE being more generalized. We also see that accuracy of all 4 methods might be similar but what they look at and what they consider important is different for all the four models. At last, we can say with confidence that combing loss functions is a very good thing for training better models.

All the experiments performed in the given report suggests that most of the deep neural networks can't be trusted fully because of their inability to generalize well. Such networks might prove very dangerous in fields like medical imaging or autonomous driving. In the future, we can look at training the networks with certain parts of the input image being hidden so it can learn more generalized results. To further understand the exact nature of the different loss functions, we can perform directional adversarial attacks instead of changing the pixel's value randomly. We can also look into training the network with only those parts where the gradient value was found relatively higher. At last, this hypothesis should be tested on a bigger dataset with scans of other body parts as well.